\documentclass[letterpaper, 10 pt, conference]{ieeeconf}  

\IEEEoverridecommandlockouts                              

\usepackage{graphics} 
\usepackage{epsfig} 
\usepackage{mathptmx} 
\usepackage{times} 
\usepackage{amsmath} 
\usepackage{amssymb}  
\usepackage{makecell}
\usepackage{booktabs}
\usepackage{multirow}
\usepackage{multicol}
\usepackage{here}
\usepackage{kotex}
\usepackage{soul, color}
\usepackage{array}
\usepackage{threeparttable}
\usepackage{caption}
\usepackage{pifont}
\newcommand{\cmark}{\ding{51}}
\newcommand{\xmark}{\ding{55}}
\DeclareMathOperator*{\argmax}{argmax}
\DeclareMathOperator*{\argmin}{argmin}
\DeclareMathOperator{\diag}{diag}
\usepackage[cal=cm,scr=euler]{mathalpha}
\usepackage{subcaption}
\usepackage{stackengine}
\usepackage{url}
\usepackage{hyperref}

\title{\LARGE \bf
GSO-SLAM: Bidirectionally Coupled Gaussian Splatting and Direct Visual Odometry
}

\author{Jiung Yeon$^{*}$, Seongbo Ha$^{*}$, and Hyeonwoo Yu$^{\dagger}$%
\thanks{*Authors contributed equally to this work. $^{\dagger}$Corresponding author.}%
\thanks{Authors are with Department of Intelligent Robotics, Sungkyunkwan University, Suwon, South Korea. \{wcr12st, sobo3607, hwyu\}@skku.edu}%
\thanks{The code is available at: \url{https://github.com/Lab-of-AI-and-Robotics/GSO-SLAM}.}%
}%

\begin{document}

\maketitle
\thispagestyle{empty}
\pagestyle{empty}

\begin{abstract}
We propose GSO-SLAM, a real-time monocular dense SLAM system that leverages Gaussian scene representation. Unlike existing methods that couple tracking and mapping with a unified scene, incurring computational costs, or loosely integrate them with well-structured tracking frameworks, introducing redundancies, our method bidirectionally couples Visual Odometry (VO) and Gaussian Splatting (GS).
Specifically, our approach formulates joint optimization within an Expectation-Maximization (EM) framework, enabling the simultaneous refinement of VO-derived semi-dense depth estimates and the GS representation without additional computational overhead. Moreover, we present Gaussian Splat Initialization, which utilizes image information, keyframe poses, and pixel associations from VO to produce close approximations to the final Gaussian scene, thereby eliminating the need for heuristic methods. Through extensive experiments, we validate the effectiveness of our method, showing that it not only operates in real time but also achieves state-of-the-art geometric/photometric fidelity of the reconstructed scene and tracking accuracy.
\end{abstract}

\section{INTRODUCTION}
Simultaneous Localization and Mapping (SLAM) has become a fundamental technique in robotics, augmented reality (AR), and virtual reality (VR), with growing demand for not only pose estimation but also geometrically accurate dense reconstruction. Therefore, dense representation SLAM methods utilizing Signed Distance Field (SDF), surfel, and other explicit representations 
\cite{newcombe2011kinectfusion, newcombe2011dtam}
have been proposed. However, with the advent of Implicit Neural Representation (INR) \cite{nerf} which models scenes as continuous functions via neural networks to enable more detailed and smoother reconstructions, INR-based SLAM \cite{orbeez-slam, imap, nice, point-slam, yang2022vox} has emerged. Nonetheless, these methods suffer from high computational costs, due to the expensive neural network inference and optimization, limiting real-time applicability.

\begin{figure}[t]
    \centering
    \includegraphics[width=\columnwidth]{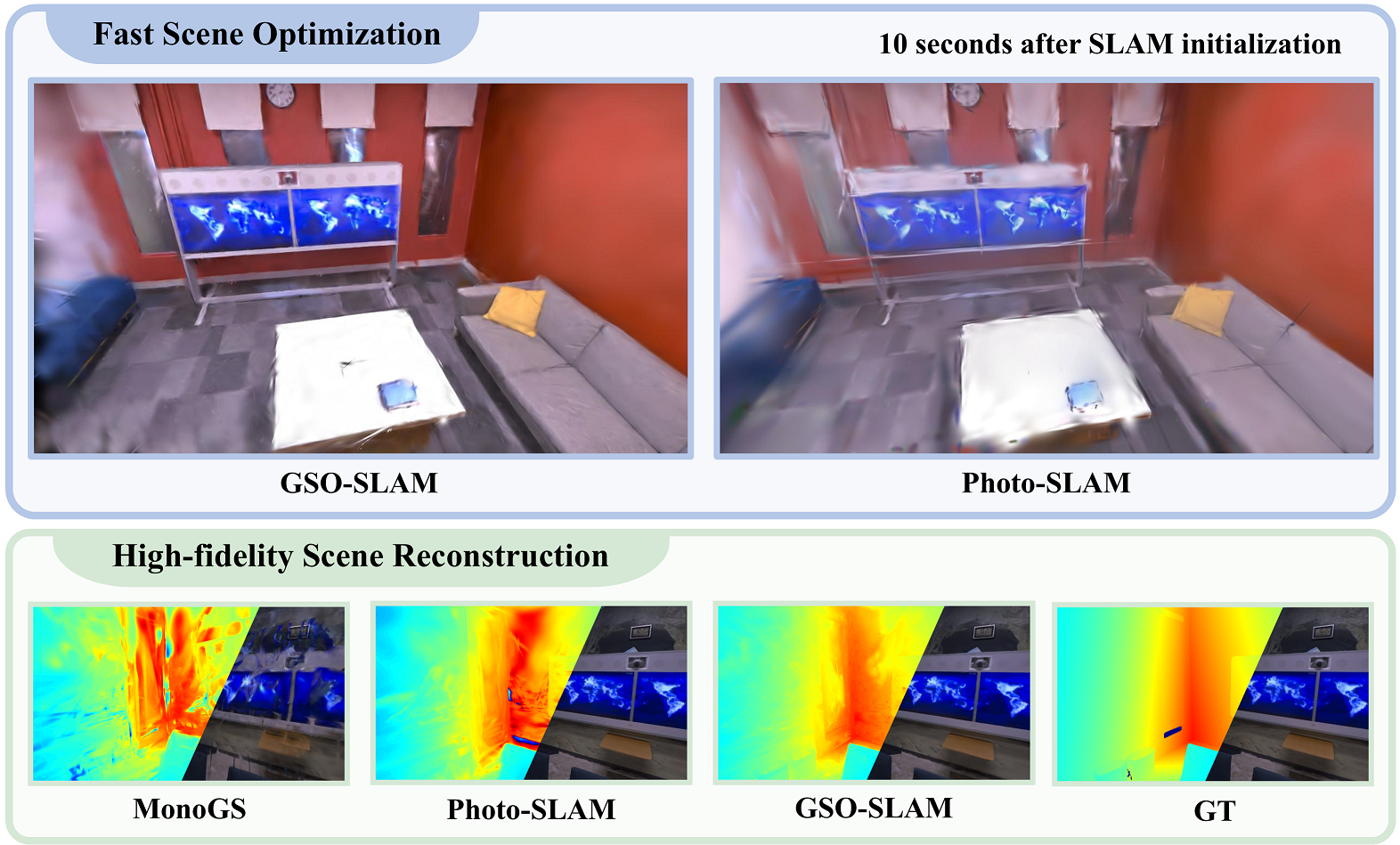}
    \caption{\textbf{Comparison of Reconstructed Scene and Rendering Results from Different SLAM Systems.} The top two images show the reconstructed 3D dense scene, highlighting the rapid reconstruction of our method. The bottom images present depth and RGB renderings, demonstrating our system's geometric accuracy and photometric fidelity.}
    \label{fig:title_figure}
    \vspace{-15pt}
\end{figure}

Therefore, Gaussian Splatting (GS) \cite{3dgs} has been introduced, which models scenes with explicit Gaussian primitives and leveraging GPU-accelerated, rasterization-based rendering.
Consequently, SLAM methods adopting GS for the spatial representation \cite{monogs, photo-slam, splatam, gs-slam, gs-icp} have been devised, showing the effectiveness of GS in dense SLAM. Among these, the coupled methods that share a unified scene for tracking and mapping \cite{monogs, splatam, gs-slam} have the advantage of directly exploiting the dense reconstructed scene by applying tracking based on re-rendering loss. However, because tracking requires repetitive rendering and optimization computations, real-time processing remains challenging even with the fast speed of GS.

To address this issue, loosely coupled approaches \cite{orbeez-slam, photo-slam, mgso} that integrate established tracking frameworks \cite{orb3, dso} with GS representation have been proposed. These methods alleviate the speed limitations of the coupled methods by performing tracking in a separate framework. Nonetheless, since the system executes tracking and mapping independently, it cannot fully exploit the reconstructed dense scene during tracking. Moreover, the integration of scene geometry from the tracking remains limited during scene optimization.

Thus, we propose a GS-based dense representation SLAM that achieves a bidirectionally coupled optimization between tracking and mapping, without additional computational overhead. Unlike prior methods that loosely integrate visual SLAM with GS, our approach formulates the optimization within an Expectation-Maximization (EM), allowing seamless interaction between Direct Sparse Odometry (DSO) \cite{dso} and GS-based mapping. Specifically, we leverage both the semi-dense depth estimates from DSO and the rendered depth from GS as pseudo-observations, enabling a tightly coupled optimization process that enhances both tracking accuracy and reconstruction quality. Importantly, this is achieved without redundant computations, as the optimization leverages the depth values computed during SLAM.

To further strengthen our coupled optimization, we introduce a novel Gaussian Splat Initialization strategy that fully exploits the existing outputs of DSO, eliminating the need for heuristic initialization methods. Since DSO computes image gradients as part of its optimization process, we directly utilize these gradients as a probabilistic estimate of the projected Gaussian Splat distribution in the image plane, deriving initial Gaussian parameters without additional computations. By jointly analyzing 2D covariances from multiple keyframes, we estimate the 3D covariance structure for new primitives, ensuring they start closer to the final converged state. This significantly accelerates the convergence of the GS scene representation while improving mapping fidelity, surpassing conventional initialization methods \cite{3dgs}.

To summarize, our contributions are as follows:
\begin{itemize}
  \setlength{\itemsep}{0pt}
  \setlength{\topsep}{1pt}
    \item We propose the first bidirectionally coupled monocular SLAM that seamlessly integrates Visual Odometry with GS, achieving real-time dense scene reconstruction with state-of-the-art photometric and geometric fidelity.
    \item Unlike previous GS-based SLAM methods that loosely couple tracking and mapping, we formulate the optimization as an EM problem, enabling joint optimization without additional computational cost. This yields improved tracking accuracy and mapping quality.
    \item We present a novel Gaussian Splat Initialization that directly exploits image intensity and gradient computed within DSO, thereby accelerating convergence and enhancing reconstruction fidelity.
\end{itemize}

\section{Related Work}
\noindent\textbf{INR-based dense SLAM} has leveraged INR \cite{nerf} to achieve high-fidelity spatial mapping. iMAP \cite{imap} explores the joint optimization of tracking and mapping using implicit neural representations, while methods like NICE-SLAM \cite{nice}, Co-SLAM \cite{coslam}, and Point-SLAM \cite{point-slam} refine this approach by incorporating hierarchical multi-feature grids, multi-resolution hash grids or neural point clouds to enhance 3D reconstruction and scalability. In parallel, approaches such as iSDF \cite{isdf} have focused on efficiently computing SDF, whereas Vox-Fusion \cite{yang2022vox} has further advanced map scalability through octree expansions. To complement the tracking speed and robustness of these integrated techniques, decoupled framework \cite{orbeez-slam} separates mapping from tracking by integrating well-structured visual SLAM system \cite{orb3}.

\noindent\textbf{GS-based dense SLAM} focuses on achieving both high map quality and fast scene reconstruction through Gaussian-based representations. 
MonoGS \cite{monogs} is a GS-based dense representation SLAM system designed to handle both RGB and RGB-D inputs, employing isotropic regularization. SplaTAM \cite{splatam} introduces a silhouette filter to address under-reconstructed regions, thereby enabling robust tracking and efficient map expansion, while GS-SLAM \cite{gs-slam} adopts a coarse-to-fine tracking approach for robust camera pose estimation. Gaussian-SLAM \cite{gaussian-slam} manages active and inactive Gaussian sub-maps to ensure high-quality photorealistic rendering in large environments. CGS-SLAM \cite{csg-slam} reduces redundant Gaussians and parameters through sliding window-based masking and geometry codebook-based quantization.

Unlike the aforementioned methods that share a unified scene for tracking and mapping, Photo-SLAM \cite{photo-slam} and MGSO \cite{mgso} adopt independent tracking modules \cite{orb3, dso}, enabling real-time system performance. GS-ICP SLAM \cite{gs-icp} utilizes G-ICP \cite{segal2009generalized} for tracking and improves performance by reducing redundant computations via mutual sharing of covariance information between G-ICP and GS scene.

\section{Method}
\subsection{Preliminaries}
\noindent\textbf{Direct Sparse Odometry} \\
Visual SLAM \cite{newcombe2011dtam, lsd-slam} generally defines the task of estimating the optimal scene geometry $D^*$ and camera poses $P^*$ from the observed image sequence $I$ as follows:
\begin{equation}
    P^*, D^* = \argmax_{P, D}\ p(I \mid P, D)
\end{equation}
To leverage photometric consistency between images, DSO \cite{dso} defines a photometric loss in the form of a weighted sum-of-squared differences (SSD):
\begin{equation} \label{eqn:dso_loss}
    E_{\mathbf{p}j}
    =
    \sum_{\mathbf{p} \in \mathcal{N}_p}
    \left\| 
            (I_j[\mathbf{p}'] - b_j) -  
            \frac{t_j e^{a_j}}{t_i e^{a_i}} (I_i[\mathbf{p}] - b_i)
        \right\|,
\end{equation}
where \( i \) and \( j \) denote reference and target frames, \( \mathcal{N}_p \) is the set of pixels in \( I_i \), \( \mathbf{p}' \) is the reprojected pixel of \(\mathbf{p}\) in \( I_j \), \(a\) and \(b\) are brightness parameters, \( t \) is the exposure time.\\
In the overall system, camera poses $P$ and scene geometry $D$ are optimized by minimizing the total photometric loss \(E_{photo}\) through a sliding window-based Bundle Adjustment (BA). Within the sliding window, for each keyframe \(i\) in the keyframe set \(\mathcal{F}\), the set of pixels of interest \(\mathcal{P}_i\) and the set of other keyframes in which these pixels are observed, \(\mathrm{obs}(\mathbf{p})\), are utilized to compute the total photometric loss:
\begin{align} \label{eqn:dso_total}
    E_{photo} := \sum_{i \in \mathcal{F}} \sum_{\mathbf{p} \in \mathcal{P}_i} \sum_{j \in obs(\mathbf{p})} E_{\mathbf{p}j}
\end{align}

\noindent\textbf{2D Gaussian Splatting} \\
Estimating a dense geometry from images is generally regarded as an ill-posed problem, requiring a stronger model of scene geometry. 
2D Gaussian Splatting (2DGS) \cite{2dgs} serves as an effective alternative representation for capturing such dense geometric information. Therefore, we adopt a latent variable $\mathcal{G}$ to model the scene via 2D Gaussians.

Each 2D Gaussian $g_i \subset \mathcal{G}$ is characterized by properties including mean $\boldsymbol{\mu}_i$, covariance $\Sigma_i$, opacity $\alpha_i$, and color $c_i$.
To reduce computational complexity in both rendering and optimization, we omit spherical harmonics.
From a given camera pose, 2DGS renders an image using a rasterization that determines each Gaussian’s contribution to the final pixels. By iteratively adjusting parameters to minimize the discrepancy with the observed images, 2DGS scene converges to an accurate scene representation.

A major advantage of 2DGS over 3DGS \cite{3dgs} is the rendering approach. While 3DGS evaluates Gaussian values at the pixel-ray intersection in 3D \cite{3dgslimit1}, 2DGS uses a ray-splat intersection in 2D, yielding multi-view consistent depth and better geometric accuracy.
Notably, its geometrically accurate scene reconstruction capability has proven useful in SLAM \cite{pings} and surface reconstruction \cite{wang2024space,feng2024rogs}.

\subsection{Joint Optimization} \label{section:joint_opt}
During DSO tracking, keyframe poses and the semi-dense map are optimized via BA, while 2DGS refines scene Gaussians by minimizing photometric error.
In a naive coupling, BA-derived semi-dense maps initialize the GS scene, and keyframe poses guide 2DGS optimization. However, since both processes run independently, they fail to share geometric information, leading to suboptimal tracking and reconstruction. Moreover, because BA and 2DGS optimize separate scene representations, redundant optimization steps lead to computational overhead.
To mitigate this, we bidirectionally couple both optimization processes, enabling DSO tracking and dense reconstruction to interact seamlessly. This integration enhances scene quality and tracking performance by jointly correcting errors within a unified framework, preventing isolated error propagation.

Nevertheless, jointly optimizing the Gaussians $\mathcal{G}$, camera poses $P$, and depth $D$ introduces a highly nonlinear and computationally demanding problem.
To address these challenges and ensure stable convergence, we employ the EM algorithm.

\noindent\textbf{EM formulation} 
Let $I$ denote the observed images, and $P$ and $D$ represent the camera poses and depth maps of keyframes, respectively. $\mathcal{G}$ denotes the set of 2D Gaussians that model the scene.
The complete-data likelihood can be written as $p(I,\mathcal{G}|P,D)$. For algebraic convenience, we instead express the joint density as $p(I,D,\mathcal{G}|P) = p(I,\mathcal{G}|D,P)p(D|P)$.
Here, $D$ can be regarded as a pseudo-observation: once $\mathcal{G}$ and $P$ are specified, the depth values are deterministically implied. Since we do not explicitly model $p(D\mid P)$, we assign it a non-informative (uniform) prior. This term is therefore constant with respect to the optimization variables in each EM sub-step, and its inclusion does not affect the maximization over $(D,P)$.
The complete-data log-likelihood that the EM algorithm seeks to maximize is thus given by:
\begin{equation}
    \log p (I,D,\mathcal{G}|P) 
\end{equation}
\noindent\textbf{E-step}
 In E-step, the Gaussians $\mathcal{G}$ are updated while keeping the previously estimated $P$ and $D$ fixed.
\begin{equation}
    L(q, P ,D ) = \mathbb{E}_{q(\mathcal{G})}
                \left[ \log {p(I,D,\mathcal{G}|P)} \right]
\end{equation}
$q(\mathcal{G})$ represents an approximate posterior distribution of the latent variable $\mathcal{G}$, given by $p(\mathcal{G}|I,P,D)$.
If $q(\mathcal{G})$ is assumed to be approximated by a $\delta$-function, i.e., $q(\mathcal{G}) \approx \delta ( \mathcal{G} - \mathcal{G}^* )$, E-step can be reformulated as a Maximum A Posteriori (MAP) estimation problem for $\mathcal{G}$. Expanding the MAP estimation problem, we obtain:
\begin{align}
    \nonumber
    \mathcal{G}^* = \argmax_\mathcal{G} &\log p(\mathcal{G}|I,D,P) \\ \nonumber
        \propto \argmax_\mathcal{G} &\log p(I|\mathcal{G},D,P) p(D|\mathcal{G},P) p(\mathcal{G}|P) \\ \nonumber
        = \argmin_\mathcal{G} &\underbrace{-\log p(I|\mathcal{G},D,P)}_\text{RGB Rendering Loss} 
                    \underbrace{-\log p(D|\mathcal{G},P)}_\text{Semi-dense Depth Loss} \\
                    &\underbrace{-\log p(\mathcal{G}|P)}_\text{Normal Consistency Loss}
\end{align}
RGB Rendering Loss measures the error between the rendered image and the observed image, serving as the photometric loss defined in the original GS method.
We define Semi-dense Depth Loss as the L1 norm between the rendered depth $D_r$ obtained from the 2D Gaussians and the current depth estimate $D$, ensuring consistency between them.
Normal Consistency Loss stands for a prior term that enforces normal vector consistency, promoting a smoother and plausible reconstruction of the scene surface.
Through these losses, the loss for E-step is formulated as follows:
\begin{align} \label{eqn:gs_loss}
    \nonumber
    \mathcal{G}^* 
    = 
    \argmin_{\mathcal{G}}
    &\ (1 - \lambda) \mathcal{L}_1(I_r, I_{gt})
    +
    \lambda\ \mathcal{L}_{D-SSIM}(I_r, I_{gt}) \\
    &+
    \lambda_d \mathcal{L}_1(D_r, D)
    +
    \lambda_n \mathcal{L}_n
\end{align}

\noindent\textbf{M-step} 
 During M-step, we fix $\mathcal{G}^*$ obtained from E-step and update $P$ and $D$. This corresponds to the following optimization problem:
\begin{align}
    \nonumber
    P^*, D^*    = \argmax_{P,D} &\log p(I,D,\mathcal{G}^*|P) \\ \nonumber
                = \argmin_{P,D} &\underbrace{-\log p(I|\mathcal{G}^*,D,P)}_\text{BA Term} 
                                \underbrace{-\log p(D|\mathcal{G}^*,P)}_\text{Depth Regularization Term} \\
                                &-\log p(\mathcal{G}^*|P)
\end{align}
BA Term accounts for photometric errors based on multi-view geometry, and once $\mathcal{G}^*$ is fixed, the term $-\log p(G^*|P)$ has minimal influence on $P$ and can therefore be neglected.
Depth Regularization Term leverages the depth information inferred from $\mathcal{G}^*$ during E-step to ensure that the updated depth $D$ remains consistent with the overall scene structure.
However, repeatedly rendering depth from $\mathcal{G}^*$ at each iteration to accommodate changing $P$ incurs significant computational overhead.
Consequently, we employ a weighted averaging approach that serves as a robust Bayesian initialization within the basin of attraction. 
This stability allows us to omit explicit regularization during the subsequent refinement, thereby facilitating the recovery of high-frequency geometric details \cite{zuo2020detailed} while minimizing computational cost.
\begin{figure*}[t]
\vspace{5pt}
    \centering
    \includegraphics[width=16.0cm]{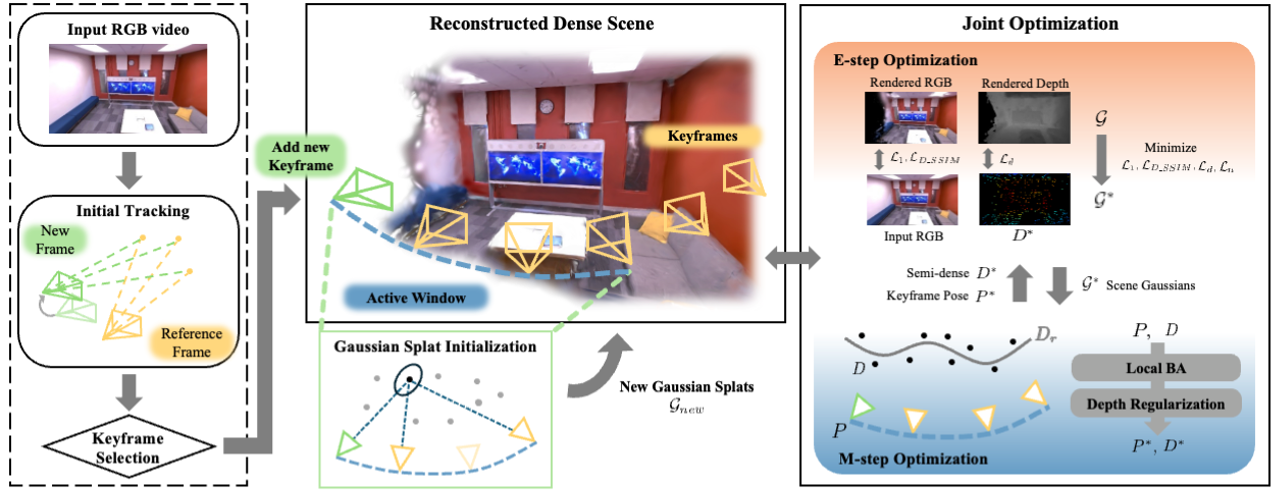}
    \caption{\textbf{SLAM System Overview.} 
    Our system reconstructs a 3D scene from monocular video. After tracking and keyframe selection, new Gaussian splats $\mathcal{G}_{new}$ are initialized, followed by EM-based joint optimization of both camera poses and the dense scene.
    }
    \label{fig:system_overview}
    \vspace{-20pt}
\end{figure*}
\vspace{-5pt}
\subsection{Gaussian Splat Initialization} \label{section:3.3}
\begin{figure}[hbt!]
\vspace{5pt}
    \centering
    \includegraphics[width=\columnwidth]{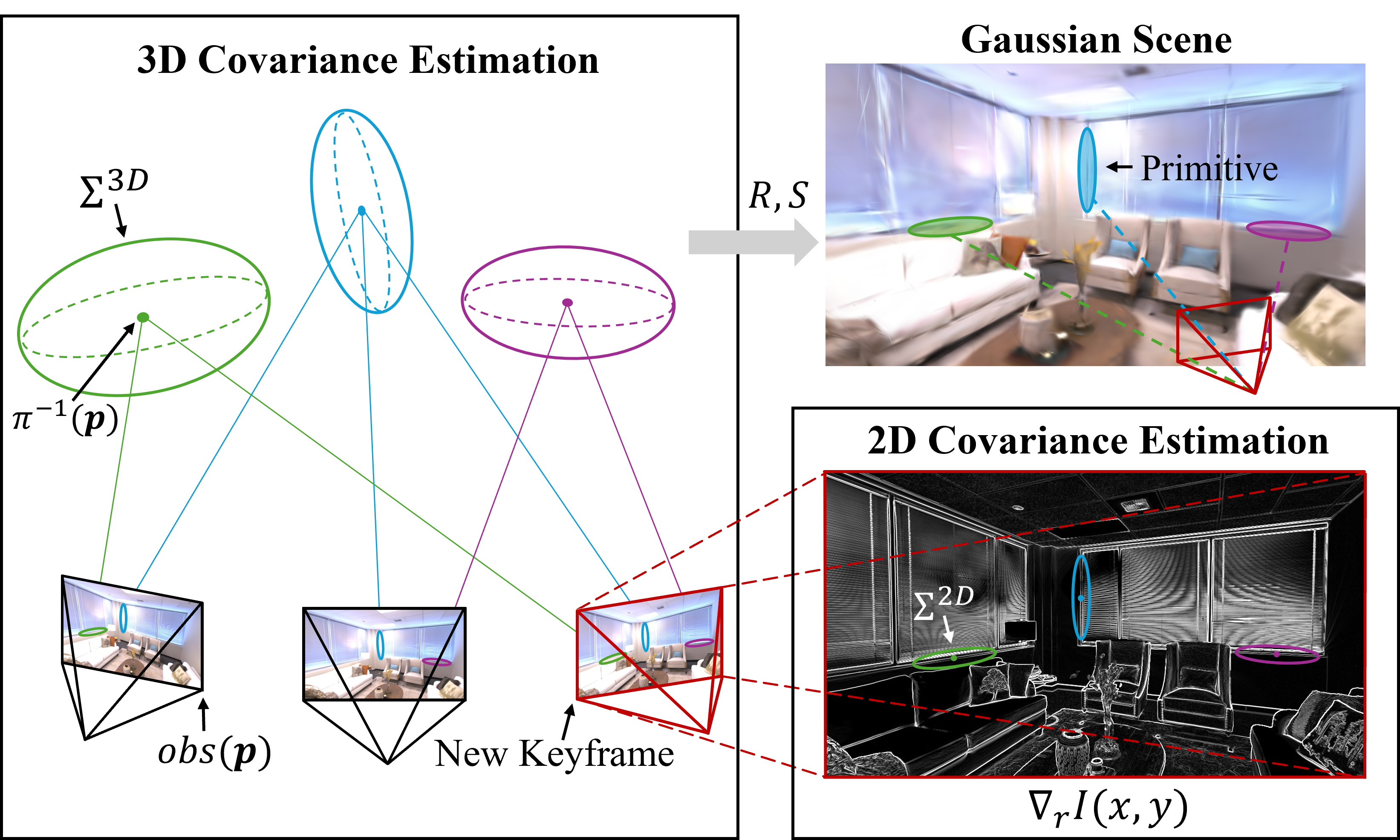}
    \caption{\textbf{Gaussian Splat Initialization.} Our initialization comprises 3 steps: (1) estimating 2D covariances from keyframe image intensities and gradients, (2) combining them to compute the 3D covariance, and (3) applying eigen-decomposition to extract the rotation and scaling parameters for the initial Gaussian.}
    \label{fig:primitive_init}
    \vspace{-20pt}
\end{figure}
Since DSO computes image gradients during the optimization process, we use these gradients as a probabilistic estimate of the projected Gaussian Splat distribution in the image plane to obtain the initial Gaussian parameters without additional computations. To this end, we assumed the intensity of the images follows Gaussian distribution \(I(x, y)\). Then, using the image coordinates from multiple keyframes obtained from DSO’s semi-dense depth estimates as centers, we compute the corresponding 2D covariances $\Sigma^{2D}$. By consolidating these, we obtain a 3D covariance $\Sigma^{3D}$ that is employed to determine the initial Gaussian parameters as shown in Fig.~\ref{fig:primitive_init}. Here, the image intensity distribution is expressed as follows:
\begin{equation}
\resizebox{\columnwidth}{!}{$
I(x, y) = \frac{1}{2\pi |\Sigma^{2D}|^{1/2}} \exp\left(-\frac{1}{2} (\mathbf{r} - \mathbf{p})^\top (\Sigma^{2D})^{-1} (\mathbf{r} - \mathbf{p}) \right),
$}
\end{equation}
where $\mathbf{r} = [x, y]^\top$, and $\mathbf{p} = [p_x, p_y]^\top$ is the center of the distribution. Subsequently, we derive the log-probability density function of the image distribution and compute its gradient to form a system of equations for various \(\mathbf{r}\). By solving this system by least-squares, we estimate \(\Sigma^{2D}\):
\begin{equation}
\nabla_{\mathbf{r}} \log I(x, y) = - \alpha (\Sigma^{2D})^{-1} (\mathbf{r} - \mathbf{p}),
\end{equation}
where \(\alpha\) is a scaling factor.

Since the same 3D point of the depth map can be observed from multiple keyframes, \(\mathrm{obs}(\mathbf{p})\), we obtain multiple 2D covariance matrices \(\Sigma^{2D}_i\) corresponding to a single 3D covariance matrix \(\Sigma^{3D}\) centered at the point. Thus, we relate the 3D covariance and the 2D covariances \cite{zwicker2001ewa} obtained from different keyframes \(\mathrm{obs}(\mathbf{p})\) as follows:
\begin{equation}
\Sigma^{2D}_i = J_i W_i \Sigma^{3D} W_i^\top J_i^\top, \quad i \in \mathrm{obs}(\mathbf{p}),
\end{equation}
where \(J\) is Jacobian matrix relating the 3D point to the image plane, and \(W\) denotes the transformation from the world to camera coordinates. To estimate \(\Sigma^{3D}\), we linearize and stack the equations from all observations into least-squares.

However, since the closed-form solutions for the 2D and 3D covariance matrices as described can result in matrices that do not satisfy the essential properties of covariance, we employ correction strategies as follows:
\begin{equation}
\tilde{\Sigma}^{3D} =
\begin{cases} 
\Sigma^{3D} + \epsilon \mathrm{I}, & \text{(Regularization)} \\ 
U \tilde{\Lambda} U^\top, & \text{(Eigenvalue clipping)}
\end{cases}
\end{equation}
where \(\epsilon>0\) is a small constant, \(U\) contains the eigenvectors of \(\Sigma^{3D}\),
\(\tilde{\Lambda}=\diag(\max(\lambda_i,\epsilon))\) with \(\lambda_i\) the eigenvalues, and \(\mathrm{I}\) denotes the identity matrix.

The 3D covariance matrices obtained from above are decomposed via eigen-decomposition. We use the eigenvector matrix as the rotation matrix $R$ and the diagonal matrix of eigenvalues as the scaling matrix $S$ by setting its smallest element to zero, yielding the initial Gaussians \cite{2dgs}. Here, we define \(\alpha\) as a constant determined by dividing a preset value by the largest element in $S$ to regularize it.
\subsection{SLAM Framework}
In this section, we present the components and algorithms of the proposed SLAM framework. The overall system architecture is shown in Fig.~\ref{fig:system_overview}.
\subsubsection{Localization and M-step Optimization}
When a new frame arrives, the SLAM system estimates the frame's pose by optimizing the energy function in Eq.~\eqref{eqn:dso_loss} using conventional two-frame direct image alignment.
In this process, the system uses the previous keyframe as the reference frame.


Next, based on the estimated pose, the system evaluates changes in field of view, translation, and camera exposure time to decide whether to select the current frame as a keyframe.
If the current frame is selected as a new keyframe, it is incorporated into the sliding window, and local BA is performed to optimize the camera pose, affine brightness coefficients, inverse depth, and camera intrinsics.
This optimization corresponds to M-step defined in~\ref{section:joint_opt}.
For depth regularization, the sparse depth maps stored in the keyframes are combined with the rendered depth map from the 2DGS scene, updated in the previous E-step. 
Their average serves as the initial depth estimate, effectively leveraging the benefits of dense reconstruction while preserving the reliability of DSO-derived depth.
Finally, we refine both the camera pose and the semi-dense depth map of the keyframes by applying the optimization in Eq.~\eqref{eqn:dso_total}.
\subsubsection{E-step Optimization}

While the main thread handles front-end processes such as tracking and keyframe selection, the 2DGS scene optimization runs in a parallel thread. 
Within our EM framework, this optimization corresponds to E-step.

Whenever the front-end selects a new keyframe, we expand the reconstructed scene by incorporating the captured scene geometry. First, the initialization in~\ref{section:3.3} defines the initial shape of the new Gaussian splats, which we then insert into the reconstructed scene. Although this yields sufficiently dense geometry, we further employ the densification approach from the original GS to improve finer details.

During parallel scene optimization, we prioritize recently added keyframes for windowed refinement and sample additional keyframes to incorporate diverse viewpoints once the active window becomes well-trained.
We run this optimization in a separate thread to ensure the front-end maintains real-time performance while the E-step continuously refines the scene Gaussians in the background.
\section{Experiments}
This section provides details on the implementation and evaluates the proposed system by comparing it with state-of-the-art approaches.
In addition, we validate the effectiveness of each proposed component by conducting ablation studies.

\noindent\textbf{Implementation Details}
Our system is implemented in C++, while the 2DGS is built using LibTorch library.
In the 2DGS optimization stage, we set the loss hyperparameters in Eq.~\eqref{eqn:gs_loss} to $\lambda = 0.2$, $\lambda_d = 500$, $\lambda_n = 0.1$.
All experiments were conducted on a desktop with Ryzen 5900X CPU, 64GB RAM, and NVIDIA RTX 4090 GPU.

\noindent\textbf{Datasets and Evaluation Metrics.}
We validate our method on synthetic Replica~\cite{replica_dataset} and real-world TUM-RGBD~\cite{tum_dataset} datasets. 
To demonstrate scalability, we additionally evaluate on the large-scale INS dataset~\cite{mneslam}, along with self-captured sequences from a quadrupedal robot (Fig.~\ref{fig:rendering_real}). 
We report ATE RMSE for tracking, and standard photometric (PSNR, SSIM, LPIPS) and geometric (L1 depth) metrics for reconstruction.
\noindent\textbf{Baseline Methods.}
We compare our method with both INR-based \cite{orbeez-slam, point-slam} and GS-based \cite{monogs, photo-slam, splatam} dense representation SLAM approaches to comprehensively evaluate its effectiveness.
The implementation of MGSO \cite{mgso} is not yet publicly available, so we reimplemented it based on the details provided in the paper and used it for our evaluation. A slight difference from the original implementation is that our version is based on 2DGS and can be described as a simple combination of DSO and 2DGS, lacking joint optimization and Gaussian Splat Initialization, unlike our method. All reported results are averages over three runs.



\begin{table}[hbt!]
  \caption{\textbf{Camera Tracking Accuracy on Replica Dataset for Monocular and RGB-D (ATE RMSE [cm] $\downarrow$). } MGSO* refers to our reimplementation of MGSO. Our method surpasses all other RGB-based systems in most scenes and achieves accuracy comparable to RGB-D-based methods.
  Note that RGB-D methods generally outperform monocular methods due to the availability of dense depth priors. Best overall results are highlighted in bold.
  }
  \centering
  { \huge
      \resizebox{\columnwidth}{!}{
        \begin{tabular}{lccccccccc}
          \toprule
          Method     & r0 & r1 & r2 & o0 & o1 & o2 & o3 & o4 & Avg. \\ 
          \midrule
          \multicolumn{10}{l}{\textit{\textbf{RGB-D Methods}}} \\
          \midrule
          Point-SLAM \cite{point-slam} &0.55 &0.41 &0.38 & 0.33 &0.51 &0.49 &0.56 &0.69 & 0.49 \\
          SplaTAM \cite{splatam}       & 0.31 & 0.39 & 0.28 & 0.49 & 0.23 & 0.30 & 0.33 & 0.60 & 0.37 \\
          MonoGS \cite{monogs}         &0.35 & 0.21 & 0.29 &0.35 & 0.21 & \textbf{0.26} & 0.13 &0.79 & \textbf{0.32} \\
          \midrule
          \multicolumn{10}{l}{\textit{\textbf{Monocular Methods}}}\\
          \midrule
          Orbeez-SLAM \cite{orbeez-slam} & 0.31 & 0.88 & 0.41 & 0.39 & 6.30 & 8.53 & 0.34 & 2.72 & 2.48 \\
          Photo-SLAM \cite{photo-slam} & 0.33 & 0.55 & 0.21 & 1.14 & 0.44 & 4.51 & 0.42 & \textbf{0.59} & 1.03 \\
          MonoGS \cite{monogs}     & 5.46 & 10.05 &5.09 &15.78 &11.18 &16.73 &5.34 &66.59 &17.03 \\
          MGSO* \cite{mgso}         & 0.05 & \textbf{0.05} & 0.03 & 0.05 & 0.04 & 4.33 & \textbf{0.04} & 1.27 & 0.73 \\
          \textbf{Ours} & \textbf{0.03} & \textbf{0.05} & \textbf{0.03} & \textbf{0.05} & \textbf{0.04} & 2.30 & \textbf{0.04} & 1.17 & 0.46 \\
          \bottomrule
        \end{tabular}
      }
  }
  \vspace{-10pt}
  \label{tab:tracking_replica}
\end{table}

\begin{table}[hbt!]
  \caption{\textbf{Camera Tracking Accuracy on TUM-RGBD Dataset for Monocular and RGB-D (ATE RMSE [cm] $\downarrow$).} We categorize the evaluated systems based on their tracking approach into direct and feature-based methods.}
  \label{tab:tracking_tum}
  \centering
  {
      \resizebox{\columnwidth}{!}{
        \begin{tabular}{lcccccc}
          \toprule
          Method & Base & fr1/desk & fr2/xyz & fr3/office & Avg. \\ 
          \midrule
          \multicolumn{6}{l}{\textit{\textbf{RGB-D Methods}}} \\
          \midrule
          Point-SLAM \cite{point-slam} & Direct & 2.73 & 1.30 & 3.52 & 2.52 \\
          SplaTAM \cite{splatam} & Direct       & 3.30 & 1.32 & 5.12 & 3.25 \\
          MonoGS \cite{monogs} & Direct         & 1.47 & 1.39 & 1.66 & \textbf{1.02} \\
          \midrule
          \multicolumn{6}{l}{\textit{\textbf{Monocular Methods}}} \\
          \midrule
          Orbeez-SLAM \cite{orbeez-slam} & Feature & \textbf{1.23} & 0.23 & 1.71 & 1.05 \\
          Photo-SLAM \cite{photo-slam} & Feature   & 1.64 & 0.25 & \textbf{1.16} & 1.34 \\
          \cmidrule(lr){1-6} 
          MonoGS \cite{monogs} & Direct           & 3.29 & 4.22 & 3.70 & 3.73 \\
          MGSO* \cite{mgso} & Direct              & 3.91 & 0.21 & 10.12 & 4.74 \\
          \textbf{Ours} & Direct                  & 2.54 & \textbf{0.20} & 6.47 & 3.07 \\
          \bottomrule
        \end{tabular}
      }
  }

  \label{tab:tracking_tum}
  \vspace{-15pt}
\end{table}

\subsection{Tracking Accuracy}
Tab.~\ref{tab:tracking_replica} shows the evaluation results of tracking accuracy on Replica dataset. Our method outperforms previous RGB-based approaches in most scenes, achieving up to 50\% lower tracking error compared to the Photo-SLAM. Moreover, even when compared to RGB-D-based methods, the proposed method exhibits comparable performance.

Tab.~\ref{tab:tracking_tum} presents results on TUM-RGBD dataset.
Since TUM-RGBD was captured using an older-generation camera (Kinect v1), the dataset contains severe noise, which negatively affects tracking accuracy. Direct methods, which rely on pixel intensity consistency, are particularly susceptible to such noise, often leading to lower tracking performance compared to feature-based methods.
Because of this, our method shows lower tracking accuracy compared to feature-based approaches.
However, since modern cameras typically exhibit much lower noise levels, we employ a direct method that leverages raw intensity data to capture dense scene geometry, offering significant advantages for mapping. While our method can be vulnerable to extreme noise, it achieves both precise tracking and superior mapping results under normal conditions, as further evidenced by its strong performance on the noise-free Replica dataset.
Furthermore, our method demonstrates higher tracking accuracy than MGSO*, even in the challenging conditions of TUM-RGBD. This result highlights that, despite the inherent noise sensitivity of direct methods, our joint optimization strategy enhances tracking robustness, enabling more accurate pose estimation even in adverse conditions.

\begin{table}[hbt!]
  \caption{\textbf{Evaluation Results of Map Quality and System FPS on Replica Dataset.}
  Results denote averages across eight sequences.
  Our method achieves real-time performance yielding the highest map quality in most scenes. The results for MonoGS correspond to its sequential implementation.}
  \centering
    {\huge
        \resizebox{\columnwidth}{!}{%
            \begin{tabular}{lccccc}
                \toprule
                Method & PSNR [dB] $\uparrow$ & SSIM $\uparrow$ & LPIPS $\downarrow$ & Depth L1 [cm] $\downarrow$ & FPS [Hz] $\uparrow$ \\
                \midrule
                \multicolumn{6}{l}{\textit{\textbf{RGB-D Methods}}} \\
                \midrule
                Point-SLAM \cite{point-slam} & \textbf{35.55} & \textbf{0.976} & 0.118 & - & 0.49 \\
                SplaTAM \cite{splatam}       & 34.13 & 0.970 & 0.099 & - & 0.21 \\
                \midrule
                \multicolumn{6}{l}{\textit{\textbf{Monocular Methods}}} \\
                \midrule
                Photo-SLAM \cite{photo-slam} & 30.91 & 0.911 & 0.099 & 19.37 & \textbf{30} \\
                MonoGS \cite{monogs}         & 32.01 & 0.919 & 0.171 & 25.36 & 0.84 \\
                MGSO* \cite{mgso}            & 28.17 & 0.855 & 0.204 & 11.54 & \textbf{30} \\
                \textbf{Ours}                & 34.48 & 0.943 & \textbf{0.060} & \textbf{8.12} & \textbf{30} \\
                \bottomrule
            \end{tabular}%
        }
    }
  \label{tab:mapquality_replica}
  \vspace{-10pt}
\end{table}

\begin{figure}[hbt!]
    \centering
    \includegraphics[width=\columnwidth]{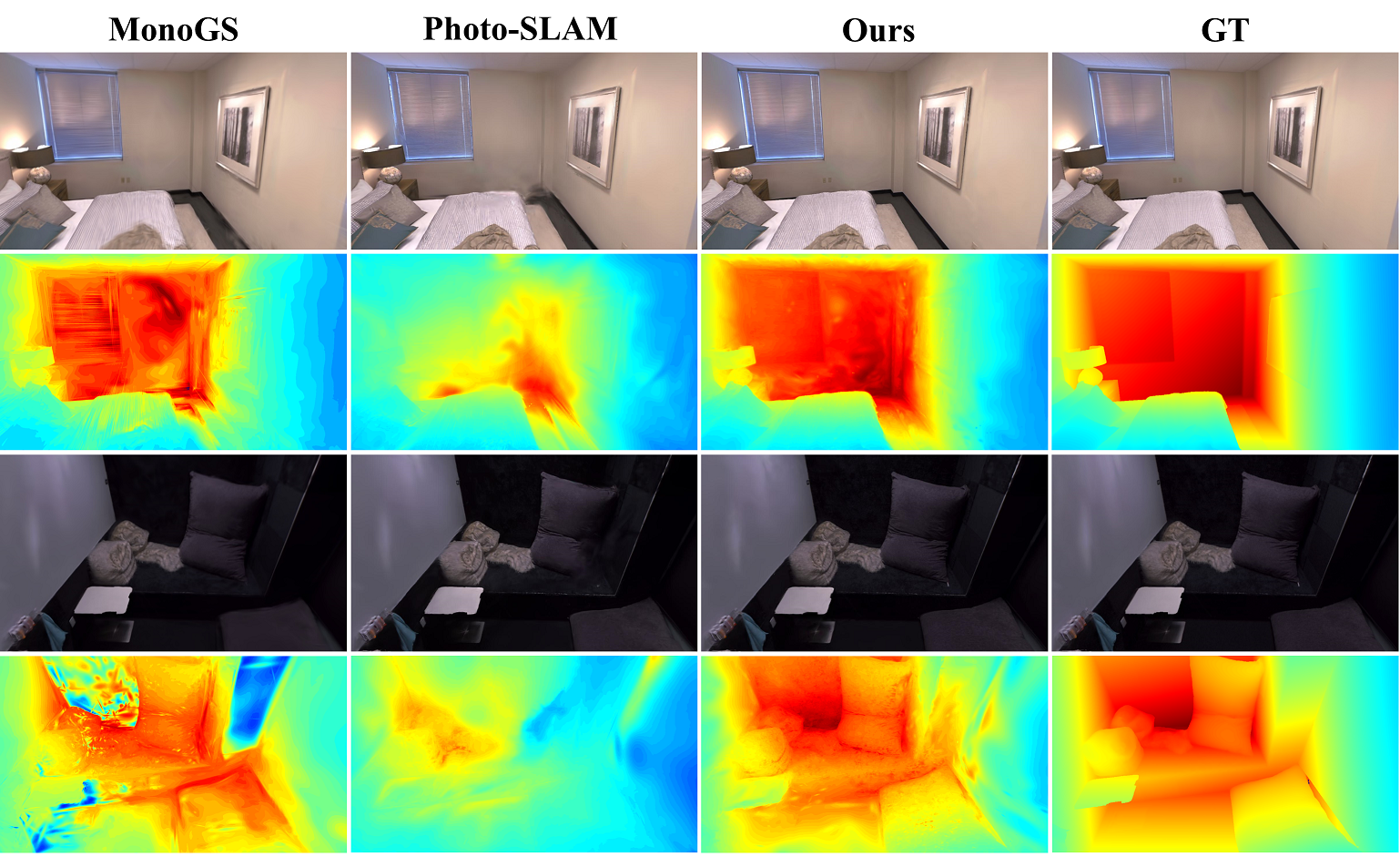}
    \caption{\textbf{Comparison of Rendering Results on Replica Dataset.} Our method exhibits superior geometric accuracy and high photometric fidelity compared to other methods.}
    \label{fig:rendering_comparison}
    \vspace{-10pt}
\end{figure}
\vspace{-5pt}
\subsection{Reconstruction Quality and System Speed}
Tab.~\ref{tab:mapquality_replica} compares reconstruction quality and system speed on Replica. We use L1 depth error to assess geometric fidelity against ground truth.
While both our system and Photo-SLAM exceed 30 FPS, we capped operation at 30 FPS to simulate realistic sensor input rates.
\begin{table}[t!]
  \caption{\textbf{Evaluation Results of Map Quality and System FPS on TUM-RGBD Dataset. } 
  The reported values represent the average across three sequences.}
  \centering
  {
    \resizebox{\columnwidth}{!}{%
    \begin{tabular}{lcccc}
      \toprule
      Method & PSNR [dB] $\uparrow$ & SSIM $\uparrow$ & LPIPS $\downarrow$ & FPS [Hz] $\uparrow$ \\
      \midrule
      \multicolumn{5}{l}{\textit{\textbf{RGB-D Methods}}} \\
      \midrule
      Point-SLAM \cite{point-slam} & 21.30 & 0.731 & 0.453 & 0.39 \\
      SplaTAM \cite{splatam}       & \textbf{23.50} & \textbf{0.909} & \textbf{0.156} & 0.41 \\
      \midrule
      \multicolumn{5}{l}{\textit{\textbf{Monocular Methods}}} \\
      \midrule
      Photo-SLAM \cite{photo-slam} & 19.78 & 0.701 & 0.232 & \textbf{30} \\
      MonoGS \cite{monogs}         & 22.15 & 0.741 & 0.321 & 3.11 \\
      MGSO* \cite{mgso}            & 19.84 & 0.690 & 0.331 & \textbf{30} \\
      \textbf{Ours}                & 20.52 & 0.727 & 0.263 & \textbf{30} \\
      \bottomrule
    \end{tabular}%
  }
  }
  \label{tab:mapquality_tum}
  \vspace{-10pt}
\end{table}
The proposed method exhibits superior photometric fidelity in most scenes, outperforming other RGB-based methods and remaining comparable even to RGB-D methods.
Notably, it delivers high-quality reconstruction with high efficiency: our system is $36\times$ faster than MonoGS and $141\times$ faster than SplaTAM. 
This efficiency stems from our Gaussian initialization which accelerates optimization. 
Furthermore, our joint optimization of photometric and geometric errors yields significantly higher geometric precision, as shown in Fig.~\ref{fig:rendering_comparison}.
Table~\ref{tab:mapquality_tum} details map quality on the TUM dataset. Despite the severe noise in TUM, our approach outperforms feature-based methods like Orbeez-SLAM \cite{orbeez-slam} and Photo-SLAM \cite{photo-slam} by extracting richer geometry. While MonoGS achieves higher quality in fr1 and fr3 due to extensive iterations, it operates $9.6\times$ slower than our method.
\begin{figure}[t!]
\vspace{5pt}
    \centering
    \includegraphics[width=\columnwidth]{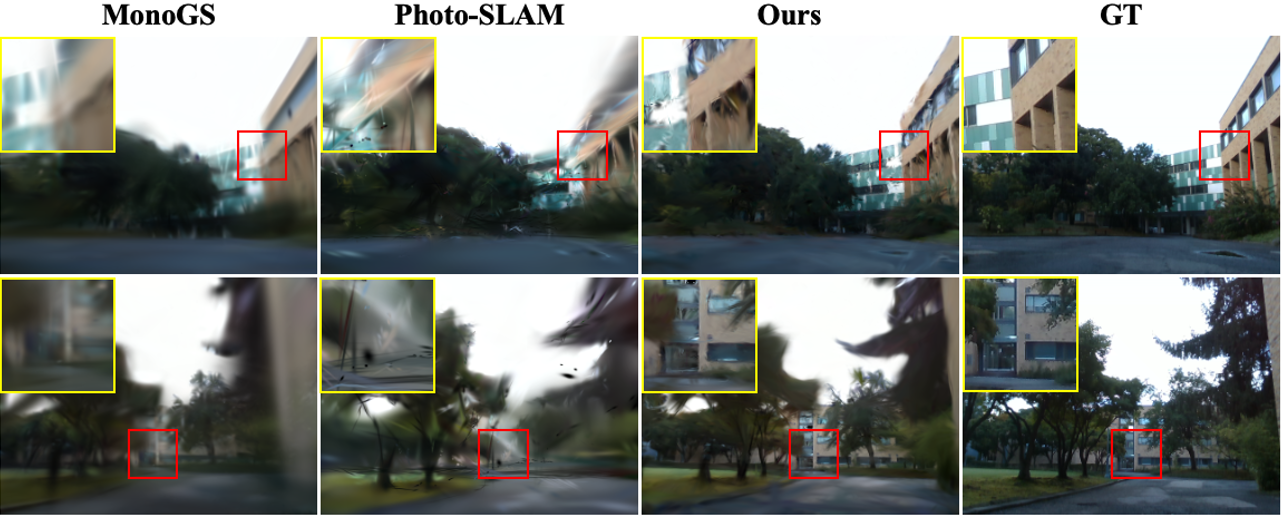}
    \caption{\textbf{Comparison of Rendering Results on Self-captured Real-world Dataset collected by a Quadrupedal Robot.} RGB images rendered from the reconstructed scenes of RGB-based approaches. Our method exhibits better visual photometric fidelity and sharper details than other methods.}
    \label{fig:rendering_real}
    \vspace{-15pt}
\end{figure}
Fig.~\ref{fig:rendering_real} illustrates the results on our self-captured dataset. Despite the significant camera motion jitter from being captured on a quadruped platform, our method achieves superior map reconstruction quality. 
\vspace{-5pt}
{
\subsection{Scalability in Large-scale Environments}
To demonstrate the scalability of our system, we conducted additional evaluations on the INS Dataset~\cite{mneslam}, which features long indoor corridors. 
\begin{table}[hbt!]
  \caption{\textbf{Tracking Accuracy Evaluation on the Large-scale INS Dataset.} 
  We report the Absolute Trajectory Error RMSE [m] on the Indoor Long Corridor (LC) sequences.}
  \label{tab:ins_tracking_accuracy}
  \centering
  \resizebox{\columnwidth}{!}{%
    \begin{tabular}{l|c|ccc|c}
      \toprule
      Method & Loop Closure & LC\_2 & LC\_3 & LC\_4 & \textbf{Avg.} \\
      \midrule
      MonoGS~\cite{monogs} & \xmark & 13.19 & 10.65 & 3.27 & 9.48 \\
      Photo-SLAM~\cite{photo-slam} & \xmark & 2.75 & 14.68 & 3.92 & 6.18 \\
      Photo-SLAM~\cite{photo-slam} & \cmark & \textbf{0.21} & 8.71 & 3.17 & 3.12 \\
      \textbf{Ours} & \xmark & 0.47 & \textbf{0.44} & \textbf{1.00} & \textbf{0.64} \\
      \bottomrule
    \end{tabular}%
  }
  \vspace{-5pt}
\end{table}
%
As Tab.~\ref{tab:ins_tracking_accuracy} shows, our method outperforms MonoGS (pure VO) in both accuracy and real-time efficiency. 
Compared to Photo-SLAM, while its global Loop Closure provides an advantage in the simple loop of LC\_2, our method proves superior accuracy in LC\_3 and LC\_4. 
In these sequences, which involve reverse-direction traversal, Photo-SLAM struggles even with loop closing enabled. This demonstrates that our core odometry minimizes drift accumulation purely through incremental tracking, surpassing baselines without relying on global backend optimization.
%
The map quality also surpasses both baselines. Averaged across all sequences, our method achieves a PSNR of 23.11 dB, significantly outperforming both MonoGS (16.76 dB) and Photo-SLAM (17.62 dB).

}
{
\begin{table}[hbt!]
    \caption{\textbf{System Runtime Analysis across Different Scales.} 
    We report the average runtime per module (in ms) on the small-scale (fr3-office) and large-scale (LC\_2) datasets.}
    \label{tab:system_speed_complexity}
    \centering
    \resizebox{\columnwidth}{!}{%
    \begin{tabular}{lccccc}
        \toprule
        Sequence & Scale & \makecell{DSO\\Tracking} & \makecell{EM Opt\\(M-step)} & \makecell{GS\\Init} & \makecell{Render\\(E-step)} \\
        \midrule
        fr3-office & Small & $1.37 \scriptstyle{\pm 2.47}$ & $50.59 \scriptstyle{\pm 9.40}$ & $5.33 \scriptstyle{\pm 1.64}$ & $1.67 \scriptstyle{\pm 0.28}$ \\
        LC\_2      & Large & $0.80 \scriptstyle{\pm 0.64}$ & $44.79 \scriptstyle{\pm 8.79}$ & $5.27 \scriptstyle{\pm 1.10}$ & $1.87 \scriptstyle{\pm 0.92}$ \\
        \bottomrule
    \end{tabular}%
    }
    \vspace{-15pt}
\end{table}
\vspace{-5pt}
\subsection{Runtime Analysis and Scalability}

Tab.~\ref{tab:system_speed_complexity} details the system runtime across small-scale (TUM-RGBD, fr3-office) and large-scale (INS, LC\_2) environments. 
Tracking maintains low latency by relying on local keyframes, while the M-step exhibits constant complexity due to our windowed-optimization strategy, which ensures bounded costs regardless of map expansion.
Runtime is influenced more by camera motion than scene scale. The rapid viewpoint changes in the handheld fr3-office sequence require more iterations for convergence compared to the smooth trajectory of LC\_2.
Regarding resource efficiency, the system recorded average CPU/GPU usage of 11.97\%/82.96\% (Small) and 9.51\%/86.17\% (Large), with peak GPU memory usage of 2.8 GB and 5.8 GB, respectively.
}
\vspace{-5pt}
\subsection{Ablation Study}
\vspace{-5pt}
\begin{table}[hbt!]
    \vspace{-5pt}
  \caption{\textbf{Ablation Study of Gaussian Splat Initialization.}
  Reported values represent the average across 8 scenes of Replica dataset.
  }
  \centering
  {
      \resizebox{0.9\columnwidth}{!}{
        \begin{tabular}{l|c|cc}
            \toprule
            \multirow{2}{*}{Method} & \multirow{2}{*}{PSNR [dB] $\uparrow$} & \multicolumn{2}{c}{Additional} \\
            & & Iterations & Time [s] \\
            \midrule
            KNN    & 25.51 & 2700 & 64.4 \\
            Const. & 31.68 & 1200 & 20.4 \\
            \midrule
            \textbf{Obs. (Proposed)} & \textbf{34.48} & \textbf{-} & \textbf{-} \\
            Pre. & 33.87 & - & - \\
            Int. & 33.92 & - & - \\
            \bottomrule
        \end{tabular}
      }
  }
  \label{tab:ablation_initialization_final_psnr}
  \vspace{-5pt}
\end{table}

\noindent\textbf{Gaussian Splat Initialization}
We validate our initialization method by two experiments.
Tab.~\ref{tab:ablation_initialization_final_psnr} presents an ablation study comparing the reconstructed scene quality under different strategies. The KNN-based initialization follows the approach used in Gaussian Splatting \cite{3dgs, 2dgs} and Photo-SLAM, while the “constant” initialization replaces KNN-based initialization with a simple isotropic Gaussian.
The obs method derives the initial Gaussian parameters described in~\ref{section:3.3}. In contrast, the pre and int methods differ in how they select keyframe \(i\) for computing $\Sigma^{2D}$ during the calculation of $\Sigma^{3D}$. Specifically, the pre method uses only the keyframe preceding the reference, while the int method integrates both the obs and pre strategies.

As shown in the table, our initialization methods achieved higher map quality than other methods, by leveraging image structural information rather than solely geometric distribution.
Among the obs, pre, and int variants, the obs method performs best. By selecting co-visible keyframes $obs(\mathbf{p})$ based on a photometric loss threshold, it computes a $\Sigma^{3D}$ that accurately reflects scene depth.
The Additional Iterations / Time in Tab.~\ref{tab:ablation_initialization_final_psnr} quantifies the acceleration, measuring the extra optimization required by baselines to match our final quality.
In Replica room1, matching our PSNR (34.45) required 2,700 additional iterations (64.6s) for KNN and 1,200 (20.4s) for constant initialization. This confirms that our strategy achieves high-quality reconstruction with significantly fewer optimization steps.

\begin{table}[hbt!]
  \caption{\textbf{Ablation of Joint Optimization.} 
  We report the average results across all scenes of Replica dataset. Our results show that the joint optimization significantly improves both the geometric fidelity and tracking accuracy.
  }
  \centering
  \resizebox{0.9\columnwidth}{!}{
    \begin{tabular}{c|ccc}
      \toprule
      Joint Opt.  & PSNR [dB] $\uparrow$ & ATE RMSE [cm] $\downarrow$ & Depth L1 [cm] $\downarrow$ \\ 
      \midrule
      \xmark    & 34.10 & 0.688 & 9.744 \\
      \cmark    & \textbf{34.48} & \textbf{0.462} & \textbf{8.124} \\
      \bottomrule
    \end{tabular}
  }
  \label{tab:ablation_sparse_depth}
  \vspace{-7pt}
\end{table}

\begin{figure}[t]
    \centering
    \setstackgap{L}{1.2\baselineskip}
    \def\sublabel#1#2{\stackinset{l}{2pt}{t}{2pt}{\color{white}\small\textbf{(#1)}}{\includegraphics[width=\linewidth]{#2}}}
    \begin{subfigure}{0.25\columnwidth}
        \centering
        \sublabel{a}{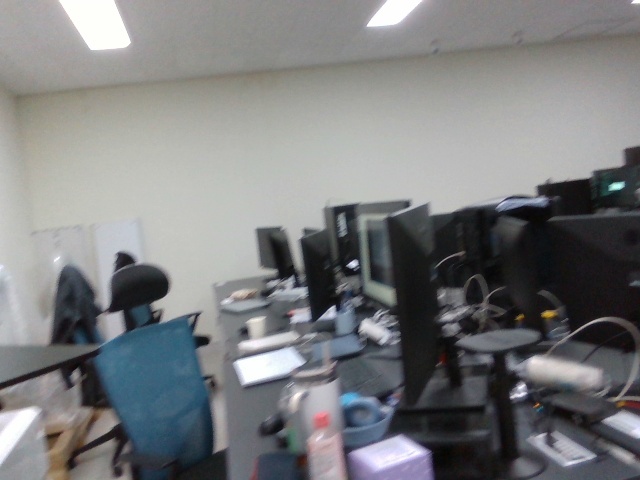}
        \phantomcaption\label{fig:blur_a}
    \end{subfigure}%
    \hfill
    \begin{subfigure}{0.25\columnwidth}
        \centering
        \sublabel{b}{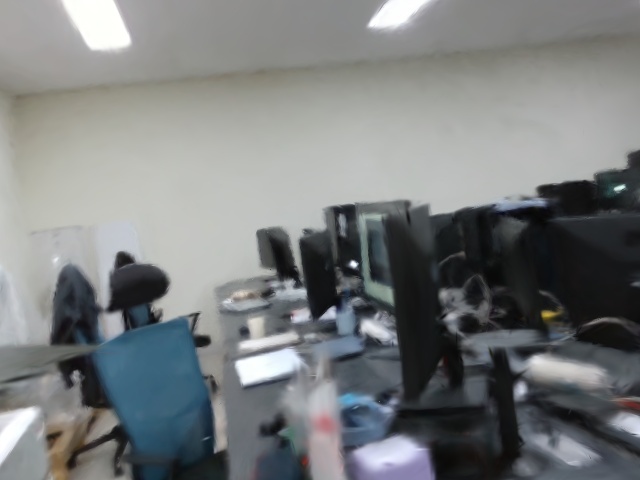}
        \phantomcaption\label{fig:blur_b}
    \end{subfigure}%
    \hfill
    \begin{subfigure}{0.25\columnwidth}
        \centering
        \sublabel{c}{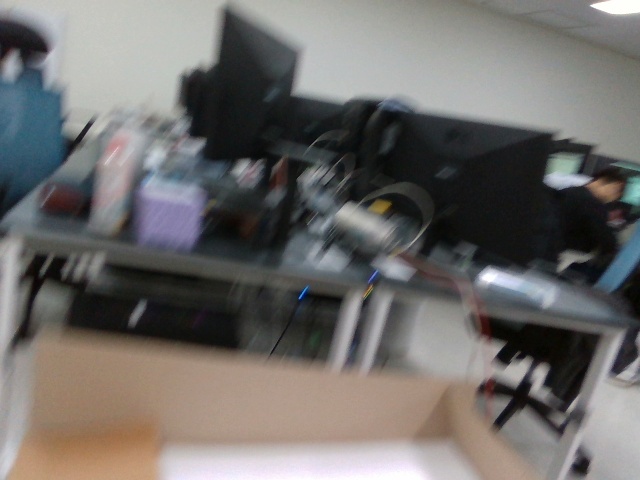}
        \phantomcaption\label{fig:blur_c}
    \end{subfigure}%
    \hfill
    \begin{subfigure}{0.25\columnwidth}
        \centering
        \sublabel{d}{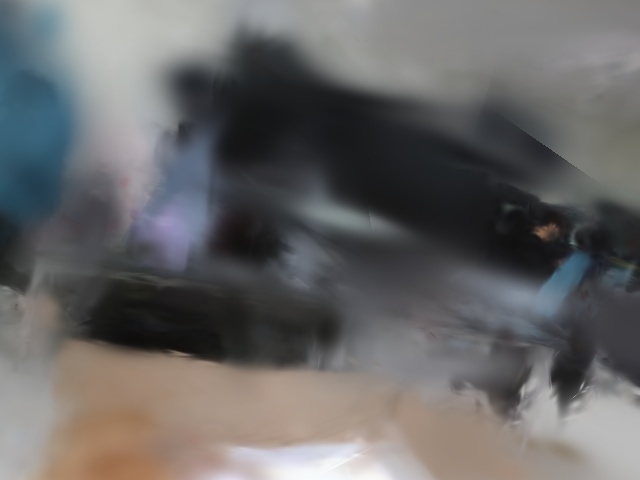}
        \phantomcaption\label{fig:blur_d}
    \end{subfigure}
    
    \vspace{-4.25mm}
    
    \begin{subfigure}{0.33333\columnwidth}
        \centering
        \sublabel{e}{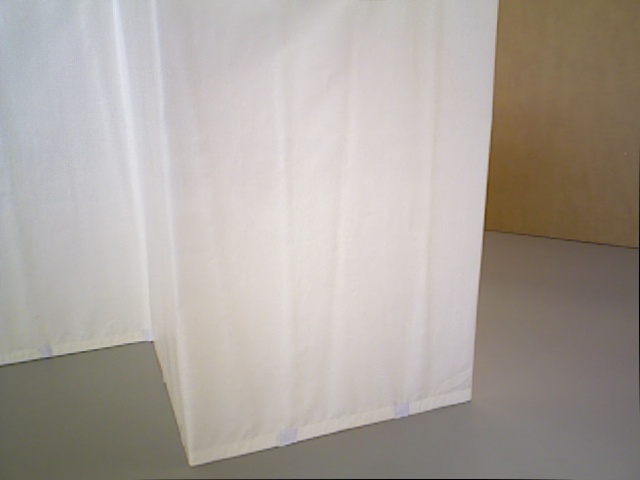}
        \phantomcaption\label{fig:text_e}
    \end{subfigure}%
    \hfill
    \begin{subfigure}{0.33333\columnwidth}
        \centering
        \sublabel{f}{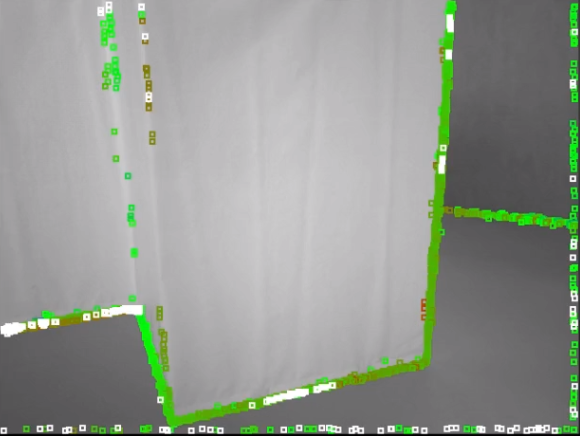}
        \phantomcaption\label{fig:text_f}
    \end{subfigure}%
    \hfill
    \begin{subfigure}{0.33333\columnwidth}
        \centering
        \sublabel{g}{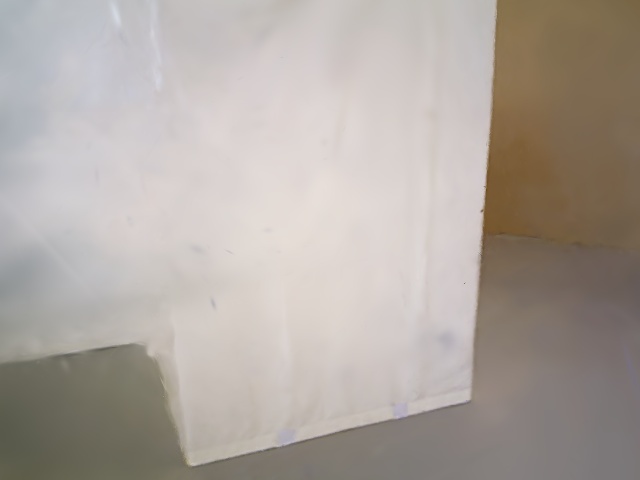}
        \phantomcaption\label{fig:text_g}
    \end{subfigure}

    \vspace{-4.0mm}

    \caption{\textbf{Qualitative Analysis of Challenging Scenarios.} 
    (a) Stable input image and (b) rendered scene reconstructed from (a). 
    (c) Input image with motion blur and (d) rendered scene reconstructed from (c). (e) Textureless input image yields (f) sparse high-gradient pixels selected by the system, resulting in (g) incomplete reconstruction in the interior regions.}
    \label{fig:failure_cases}
    \vspace{-15pt}
\end{figure}

\noindent\textbf{Joint Optimization}
%
Tab.~\ref{tab:ablation_sparse_depth} presents the results of the ablation study on the effectiveness of Joint Optimization.
The case without Joint Optimization (\xmark) corresponds to the ``naive coupling'' strategy discussed in~\ref{section:joint_opt}.
In this setting, while the GS scene is initialized using BA-derived semi-dense maps, the bidirectional EM framework is disabled. Consequently, vanilla DSO and 2DGS operate independently with no mutual influence between the two components.
This separation prevents the refined scene geometry from being effectively shared, limiting the overall optimization performance.
In contrast, when Joint Optimization is applied, the scene geometry estimated through BA is properly incorporated into the 2DGS optimization process, improving both the Depth L1 error and PSNR. Furthermore, as the BA process also benefits from the optimized scene representation, tracking performance is enhanced as well.
\vspace{-2.5pt}
{
\subsection{Failure Case in Extreme Scenarios}
We evaluate our system in extreme scenarios using self-captured sequences and the textureless TUM-RGBD dataset (fr3/structure\_notexture\_far and fr3/structure\_notexture\_near). As shown in Fig.~\ref{fig:failure_cases}, the system remains robust under stable motion ((a), (b)). However, severe motion blur ((c)) violates photometric consistency, causing tracking drift that propagates into geometric distortions ((d)).
%
Textureless regions, such as white walls ((e)), also present challenges. Since our method extracts initial scene geometry based on image gradients ((f)), uniform surfaces lack sampling, leading to incomplete geometry despite densification ((g)). Nevertheless, our initialization proves effective even in these textureless scenes. It achieves 25.51 dB PSNR, outperforming Constant (22.27 dB) and KNN (25.23 dB) methods while requiring significantly fewer iterations to converge.
}

\section{Conclusion}
\noindent 
In this paper, we introduce GSO-SLAM, the first bidirectionally coupled monocular SLAM method that seamlessly integrates Visual Odometry and Gaussian Splatting, eliminating redundant computations and enhancing both tracking and mapping performance.
We formulate the joint optimization of tracking and mapping as an Expectation-Maximization problem, allowing it to run without additional computational overhead while improving tracking accuracy and mapping quality. Furthermore, our approach exploits image gradients, keyframe poses, and pixel associations from Visual Odometry to derive initial Gaussian parameters, accelerating convergence and boosting reconstruction quality. Extensive experiments validate our method’s real-time performance, achieving state-of-the-art photometric and geometric fidelity in dense scene reconstruction along with superior tracking accuracy. In current work, we model the semi-dense map as a set of discrete points. In future work, we plan to incorporate more detailed modeling such as considering image structure to further enhance the benefits of our coupled approach.
%

\bibliographystyle{IEEEtran}
\bibliography{main}

\end{document}